\documentclass[twoside,11pt]{article}
\usepackage{jmlr2e}
\ShortHeadings{Balanced Random Survival Forests}{Afrin, Illangovan, Srivatsa, and Bukkapatnam}
\firstpageno{1}
\usepackage{amsmath,amssymb}
\usepackage{mathtools}
\usepackage{bm}
\usepackage{bbm}

\usepackage{calrsfs}
\DeclareMathAlphabet{\pazocal}{OMS}{zplm}{m}{n}

\usepackage{tabularx}
\usepackage{threeparttable}
\usepackage{booktabs}
\usepackage{footnote}

\usepackage{color}
\usepackage{rotating}
\usepackage{rotfloat}
\usepackage{algorithm,algorithmic}
\usepackage{enumerate}
\usepackage{hyperref}
\hypersetup{
	colorlinks=true,      
	urlcolor=black,
	citecolor=black,
linkcolor=black,
filecolor=black
}
\usepackage{color}
\begin{document}

\title{Balanced Random Survival Forests for Extremely Unbalanced, Right Censored Data}

\author{\name Kahkashan Afrin \email
	    afrin@tamu.edu \\
        \name Gurudev Illangovan \email 
	    ilan50\_guru@tamu.edu \\
		\addr Department of Industrial and Systems Engineering \\
	    Texas A\&M University\\
        College Station, TX 77843-3131, USA
	    \AND
        \name Sanjay S.\ Srivatsa \email sanjaysrivatsa@hotmail.com\\
        \addr Heart, Artery, and Vein Center of Fresno\\
        Fresno, CA 93722, USA
 	    \AND
         Satish T.\ S.\ Bukkapatnam\thanks{STSB is corresponding author} \email satish@tamu.edu \\
        \addr Department of Industrial and Systems Engineering \\
	    Texas A\&M University\\
        College Station, TX 77843-3131, USA}

\editor{}

\maketitle

\begin{abstract}% 
Accuracies of survival models for life expectancy prediction as well as critical-care applications are significantly compromised due to the sparsity of samples and extreme imbalance between the survival (usually, the majority) and mortality class sizes. While a recent random survival forest (RSF) model overcomes the limitations of the proportional hazard assumption, an imbalance in the data results in an underestimation (overestimation) of the hazard of the mortality (survival) classes.  A balanced random survival forests (BRSF) model, based on training the RSF model with data generated from a synthetic minority sampling scheme is presented to address this gap. Theoretical results on the effect of balancing on prediction accuracies in BRSF are reported. Benchmarking studies were conducted using five datasets with different levels of class imbalance from public repositories and an imbalanced dataset of 267 acute cardiac patients, collected at the Heart, Artery, and Vein Center of Fresno, CA. Investigations suggest that BRSF provides an improved discriminatory strength between the survival and the mortality classes. It outperformed both optimized Cox (without and with balancing) and RSF with an average reduction of 55\% in the prediction error over the next best alternative.
\end{abstract}
%\cmmt{consistent with biomarker/predictor/variable}
%\cmmt{table for individual data improvement of IBS error? }

\section{Introduction} \label{intro}
Mortality prediction, risk stratification, and key biomarker identification in acute and high-risk patients are prominent tasks in healthcare survival analysis~\citep{amsterdam20142014,o20132013}. Their importance is further accentuated while dealing with life-threatening conditions such as acute cardiovascular diseases and cancer. An accurate prediction of the adverse pathologies can allow risk-calibrated interventions for better management of the outcomes. Survival analysis models are one of the key tools which provide decision support to the physicians not only for intervention selection but also for identifying high-risk patient and for patient counseling/consent for intervention~\citep{furnary1996long,dankner2003predictors}. Nonetheless, the limited accuracy of such survival models which guide critical life-and-death decisions remains a major concern.\par
	
Currently, most of the real world applications employ the Cox proportional hazard model for mortality prediction (CPH)~\citep{harrell2015cox}. For a given covariate vector, $\bm{x}_i$ (capturing the various biomarkers in the present context), CPH reduces the estimation of hazard function at a time $t$, $\lambda_i(t)$, for an individual $i$ into a regression problem of the form $ \lambda_i(t)=\exp(\bm{\beta}^\top\bm{x})\lambda_0(t)$, where $\bm{\beta}$ is an unknown vector of regression coefficients. The popularity of CPH rises from its semi-parametric nature, which does not require any distributional assumption of the baseline hazard function, $\lambda_0(t)$, to estimate $\bm{\beta}$. Nonetheless, selecting a wrong $\lambda_0(t)$ can significantly change the result~\citep{ohno1997comparison}. Additionally, the CPH model makes certain restrictive assumptions, many of which do not hold in real life scenarios. One such assumption is a constant hazard ratio between any two observations at every time instant, $t$. It also does not take into account the missing predictors, the nonlinearity of the exponential factor, interdependence among observations, and is known to have an inherent bias and high generalization error~\citep{binder1992fitting,snedecor1989statistical,pan2008proportional}. 

Recent innovations in the sensor technologies for gathering a rich collection of clinical biomarkers, together with advent of innovative methodologies for predicting time-to-event, based on advanced machine learning techniques have opened new possibilities to overcome the limitations of CPH models~\citep{belle2011learning, ishwaran2008random}, many of which address the limitations of CPH model. One such method is random survival forest (RSF)~\citep{ishwaran2008random}. RSF is a non-parametric approach to right-censored survival analysis based on a Breiman's ensemble tree, random forests model. In Breiman's random forests, a tree is grown using $B$ independent bootstrapped samples with a different set of biomarkers at each node, randomly selected from $\bm{x}$. This two-way randomization improves both bias and variance of the resulting random forests ensemble. Its performance is at least comparable to that of the state-of-the-art machine learning methods, such as boosting and support vector machine~\citep{ishwaran2010consistency}. Additionally, RSF effectively imputes the missing data---a common problem in healthcare datasets. RSF inherits the robustness and desirable properties (increased accuracy, minimized bias, and variance) of random forests model to the survival analysis. Recent works using RSF for survival data have shown improved results as compared to the CPH models and are getting popular as a survival analysis tool~\citep{hsich2011identifying,mogensen2012evaluating}. \par	
Nonetheless, the characteristics of the survival data pose significant challenges to RSF. Besides right-censoring, the presence of extreme imbalance between the censored and the mortality classes with as low as 2-10\% data in the minority is a commonly occurring, yet often ignored aspect. Due to the contemporary clinical practice and infrastructure across the US, acute cardiac and other life-threatening diseases are mostly treated in small tertiary care hospitals and, as a result, the cohort size tends to be small, further exacerbating the challenge. Balancing is an essential step in maximizing the utility and improved mortality prediction performance. Although data balancing is important, only a few works focuses on addressing class imbalance from a survival analysis context~\citep{chia2012looking}. In this work, we propose a balanced random survival forests (BRSF), which integrates RSF with a synthetic data balancing scheme. We present some key theoretical result on the effect of class imbalance on improving model's predictive performance from a survival analysis context. The performance of BRSF was with RSF as well as, an optimized CPH model and its balanced counterpart. Here, optimized CPH refers to the CPH model where overfitting errors are minimized through predictor selection. All models are assessed on a set of 5 benchmark datasets each representing a different degree of class imbalance, as well as a dataset gathered at the Heart, Artery, and Vein center of Fresno from 267 acute cardiac STEMI (ST Elevated Myocardial Infarction) subjects after they underwent cardiac revascularization therapy. The paper reports the following three contributions, namely, the development of BRSF approach to address the challenges with high class imbalance and small data size in survival analysis, establishment of theoretical results on how data balancing can improve model prediction, and comparison of the performance of BRSF models relative to that of the other contemporary survival models in multiple scenarios with high data imbalance, which collectively can enhance informed treatment decision for healthcare providers. \par
The remainder of this paper is organized as follows. We describe the BRSF modeling approach and delineate the performance comparison measures in Section~\ref{method}. In Section~\ref{case}, we provide the details for the survival datasets used in this paper and the comparative results obtained before and after addressing the class imbalance. Finally, Section~\ref{summary} summarizes the paper and discuss some future work for effectively balancing the survival data.

\section{Balanced RSF for Mortality Prediction} \label{method}
The BRSF model consists of two main aspects, namely, the tree-based random survival forest modeling, and the class balancing methodology, as presented in the following subsections.
	
\subsection{An Overview of Random Survival Forests}\label{section 2}
\begin{figure}[!htb]
		%\centering
		\includegraphics[width = 1\textwidth]{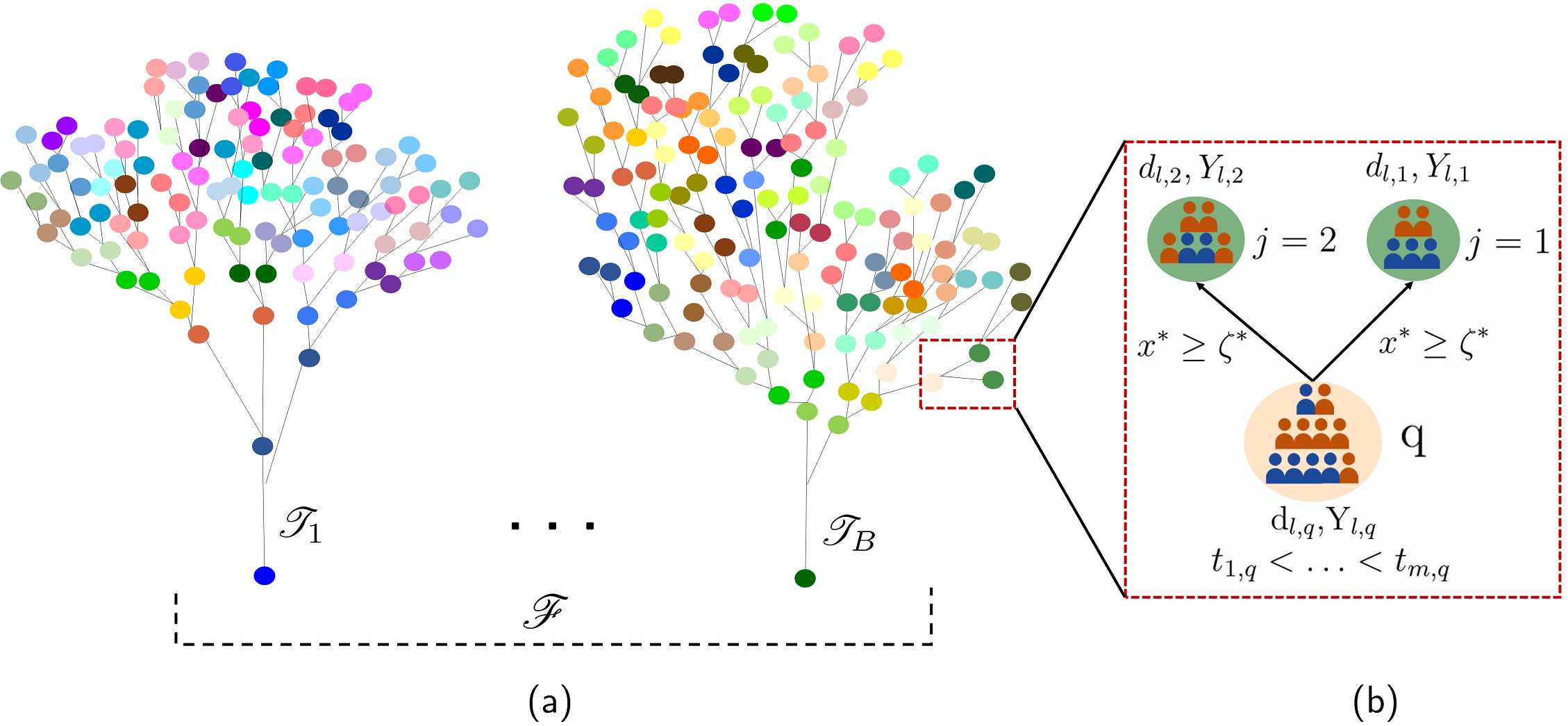}
				 %\captionsetup{justification=centering}
		\caption{ A pictorial representation of (a) an RSF ($\mathcal{F}$) consisting of $B$ trees (b) split of a parent node, $q$ into two daughter nodes using the biomarker $x^*$ at value $\zeta^*$}
		\label{RSF}
	\end{figure}

Growing a random survival forest, $\mathcal{F}$ can be thought of as a hierarchical procedure which initializes by randomly drawing $B$ bootstrap samples from the training data consisting of $N$ samples, each with $R$ predictors (here, biomarkers), and growing a survival tree $\left\{\mathcal{T}_b\right\}_{1 \leq b \leq B}$ for each of the drawn samples (see Figure~\ref{RSF}). The bootstrap samples are invariably extracted from right-censored survival data. For analyzing survival data, follow up time and associated right censoring are important considerations. Right-censored survival data of $N$ individuals is the collection of values in a set, $\bm{\Phi}_i=\left\{\left(\bm{x}_i, T_i, \delta_i\right) \right\}_{1\leq i\leq N}$,  where the subscript $i$ is the patient index, and $\bm{x}_i$ are independent and identically distributed (i.i.d.)~biomarkers of patient $i$. Let $T^0_i$ and $\pazocal{C}_i$ be the true event (death) and censoring time, respectively for subject $i$. The observed survival time is then given as $T_i=\min(T^0_i, \pazocal{C}_i)$, and $\delta_i:=\mathbbm{1}_{T_i^0\leq \pazocal{C}_i}$ is the binary censoring status specified as follows: given the vector of biomarkers, $\bm{x}_i=(x_i^r), i=1,\ldots,N; r=1,\ldots,R$, an individual $i$ is said to be right-censored if $T^0_i \leq C_i$, i.e., $\delta_i=0$ or else the individual is said to have experienced death at time at time $T_i (\delta_i=1)$.
	
Here, the construction of a survival tree, $\mathcal{T}_b$ from the $b^{th}$ bootstrapped data begins with a random selection of $p$ out of $R$ possible biomarkers in $\bm{x}$. Although we used the suggested, $p=\sqrt{R}$~\citep{ishwaran2011random, james2013introduction}, the value of $p$ depends on the number of available biomarkers and is data specific. Previous studies have even shown good performance with $p=1$, care must be taken as an increase in $p$ tend to result in correlated trees~\citep{breiman2001random}. Next, all the $N$ bootstrapped samples are assigned to the root node, i.e., the topmost node of the tree. The root node is then split into two daughter nodes, and each of thus-generated daughter nodes is then recursively split with progressively increasing within-node homogeneity. Now, for any parent node with $p$ predictors, the split on a given predictor, $x^{v}$ is of the form $x^{v} \leq \zeta^{v}_\gamma$ and $x^{v} >\zeta^{v}_\gamma; 1\leq v \leq p$. Here, $\zeta^{v}_\gamma$ conventionally takes values at the midpoint of consecutive distinct observations of $x^{v}$ corresponding to the individuals in the parent node being split~\citep{segal1988regression}. Thus, $\gamma$ has at most one less than the parent node size values.
	
Let $t_{1,q} <t_{2,q}<...<t_{m,q}$ be $m$ unique event (death) times at the parent node, $q$, and $d_{lj}$ and $Y_{lj}$ denote the number of deaths and individuals who are alive (at risk) in the daughter node $j \in \left\{1,2\right\}$ at time $\left\{t_{l,q}\right\}_{1 \leq l \leq m}$. It follows that $d_{lj}$ individuals had survival time of less than $t_{l,q}$, and $Y_{lj}$ individuals had a greater survival time. For a split using biomarker $x^{v}$ and its splitting values $\zeta^{v}_\gamma$, the goodness-of-split is measured using a log-rank statistic~\citep{segal1988regression} represented as:
\begin{eqnarray}
	L(x^{v},\zeta^{v}_\gamma)=\dfrac{\sum_{l=1}^{m}\left(d_{l,1}-Y_{l,1}\dfrac{d_{l,q}}{Y_{l,q}}\right)}
	{\sqrt{\sum_{l=1}^{m}\dfrac{Y_{l,1}}{Y_{l,q}}\left(1-\dfrac{Y_{l,1}}{Y_{l,q}}\right)\left(\dfrac{Y_{l,q}-d_{l,q}}{Y_{l,q}-1}\right)d_{l,q}}}
	\label{eq:eq1m}
	\end{eqnarray}
\noindent Here, Equation~\ref{eq:eq1m} measures the separation between two daughter nodes. Hence, the best split at a node $q$ is determined by the biomarker $x^*$ and its value at the cut point $ \zeta^*$ such that $|L(x^*,\zeta^*)|\geq |L(x^{v},\zeta^{v}_\gamma)|$ $\forall~ x^{v}~ \text{and}~ \zeta^{v}_\gamma$. Algorithm~\ref{algo} presents the procedure to select $x^*$ and $\zeta^*$ for any given parent node with $\kappa$ distinct values of $\gamma$.
	\begin{algorithm}[!htb]
		\begin{algorithmic}[1]
			\STATE Initialize:  $\left\{x^*, c^*,  L(x^*,c^*)\right\}\leftarrow 0$
			\FORALL{${v} \in \left\{1,p\right\}$}
			\FORALL{$\gamma \in \left\{1, \kappa\right\}$}
			\STATE determine $d_{l,1}; Y_{l,1}; L(x^v, \zeta^v_\gamma) $
			\IF {$|L(x^{v}, \zeta^{v}_\gamma)| > |L(x^*,\zeta^*)|$}
			\STATE $L(x^*,\zeta^*)\leftarrow L(x^{v}, \zeta^{v}_\gamma) $
			\STATE $x^* \leftarrow x^{v} $
   			\STATE $\zeta^* \leftarrow \zeta^{v}_\gamma $
			\ENDIF
			\ENDFOR
			\ENDFOR
		\end{algorithmic}
		\caption{Selecting the best split}
		\label{algo}
	\end{algorithm}

Figure~\ref{RSF} presents a pictorial representation of a RSF and the log-rank split procedure presented in Algorithm~\ref{algo}. Trees are grown until no new daughters can be formed due to a stopping criterion of a minimum of $d_0>0$ unique deaths~\citep{ishwaran2008random}. Here, we used $d_0=3$. At this point, there are $\mathcal{L}(\mathcal{T}_b)$ terminal/leaf nodes in the tree, $\mathcal{T}_b$. Now, let $h \in \mathcal{L}(\mathcal{T}_b)$ be one of the terminal nodes with $N(h)$ distinct event times, $\left\{t_{l,q}\right\}_{1 \leq l \leq N(h)}$. Following the notations introduced above, the cumulative hazard function (CHF) for the node $h$ as given in Equation~(\ref{eq:eq2m}) here, is the Nelson-Aalen estimator for the patients in $h$ at time $t$~\citep{borgan2005nelson}.
	\begin{eqnarray}
	\hat{H}_h(t)=\sum_{t_{l,h}\leq t}\dfrac{d_{l,h}}{Y_{l,h}}
	\label{eq:eq2m}
	\end{eqnarray}
\noindent For an individual $i$ with a vector of biomarkers $\bm{x_i}$, the CHF is same as that of the terminal node it belongs to, i.e. $\hat{H}(t|\bm{x}_i)=\hat{H}_h(t), ~ \text{if}~\bm{x}_i \in h$. However, due to bootstrapping (sampling with replacement) an individual can be present in several bootstrap samples and hence in several trees. Thus, to obtain an out of bag (OOB) ensemble CHF, $\hat{H}_e(t|\bm{x}_i)$ for individual $i$ at time $t$, we calculate the average of the CHFs for all such trees in which $i$ is an OOB sample. For the tree $\mathcal{T}_b$ with CHF $\hat{H}_b(t|\bm{x})$, let $I_{i,b}$ be 1 if $i$ is an OOB data for $\mathcal{T}_b$ and 0 otherwise, then the OOB ensemble CHF for that individual is given by:
	\begin{eqnarray}
	\hat{H}_e(t|\bm{x}_i)=\dfrac{\sum_{b=1}^{B}I_{i,b}\hat{H}_b(t|\bm{x}_i)}{\sum_{b=1}^{B}I_{i,b}}
	\label{eq:eq3m}
	\end{eqnarray}
In practice, hazard functions for training are calculated on the OOB data to avoid an optimistic bias in the result. For each bootstrap sample about one-third of the data is left for OOB. The main steps of the RSF algorithm are summarized in Algorithm~\ref{alg:seq1}. 
	
Now, we discuss some of the notable features of the RSF. First, introducing the two-fold randomization both by $B$ bootstrap samples as well as by $p$ candidate variables reduces the generalization error and provides RF and its related method RSF an edge over the traditional approaches. The value of $B$ can be estimated using the generalization error. Since the generalization error approaches a limiting value with increasing number of trees after a point adding more trees does not add to the increased accuracy of the forest~\citep{breiman2001random}. We selected $1000$ trees for our analysis.

Second, RSF helps in solving another key issue of missing data. As is the case for most of the real data, missing values are generally present and severely affect the error estimation. RSF uses an adaptive tree imputation method at each parent node to impute missing data (predictors and outcomes) before the node is split. For the training data at a node, $q$, imputation for the missing values of the $p^{th}$ biomarker works by randomly drawing from the empirical distribution, $\pazocal{D}_{p,q}$ of ``in-bag" non-missing set of values for $\bm{x}^p$ at node $q$. The daughter node thus formed does not contain any missing data. However, for robust results, the imputed values are reset to missing in the daughter node and the imputation is iteratively repeated at every node until the terminal node is reached. Now, the final summary imputed value for the missing biomarkers for subject $i$ is the average (across the forest) of its ``in-bag" imputed values obtained from the terminal nodes in which the subject is present. In case the biomarker is not continuous, most frequently occurring ``in-bag" imputed value is considered. Imputation for the missing values in the test data proceeds in the same way, however, the missing values are drawn from the original distribution, $\pazocal{D}_{p,q}$ from the training data~\citep{ishwaran2008random}. Third, it may be noted that with the assumption of discrete feature space, an RSF has shown to converge uniformly to the true population survival function~\citep{ishwaran2010consistency}. This demonstrates the consistency of the RSF for right-censored survival analysis. RSF indeed has several merits and might provide good predictions (measured in terms of the metrics presented in subsection~\ref{error}), but real life survival data with extreme imbalance coupled with small data sizes often make accurate prediction very challenging. We discuss the balancing methodology used to address this limitation of the RSF in subsection~\ref{balancing}.

\begin{algorithm}[!htbp]
		\begin{algorithmic}[1]
			\STATE Initialize:  $i\leftarrow 1, b \leftarrow 1, x^* \leftarrow0, \zeta^* \leftarrow 0$
			\STATE Select $B$, $d_0,\bm{\Phi}_{train} $
			\WHILE{$b \leq B$}
			\STATE Grow $\mathcal{T}_b$
			\WHILE{unique deaths in $\mathcal{L}(\mathcal{T}_b) \geq d_0$}
			\STATE Find $x^*$, $\zeta^*$
			\STATE Perform node split
			\ENDWHILE
			\ENDWHILE
			\STATE Calculate $\text{CHF}(\mathcal{F})$ for $\bm{\Phi}_{OOB}$
		\end{algorithmic}
		\caption{Growing the RSF}
		\label{alg:seq1}
	\end{algorithm}

\subsection{Performance Measures} \label{error}
In the automated prognostics and decision support practice, where data drives the critical decision-making, robustness of the model is of utmost importance. Recently, there have been vigorous debates on the effectiveness of the performance measures and on the efficacy of one measure over the other~\citep{ishwaran2011random}. Here, we compare the performance of BRSF relative to contemporary survival models based on three of the most popular metrics in survival analysis literature. These are concordance index, prediction error curves, and Integrated Brier score. Further, an accurate estimation of prediction error with limited data is a challenging task, therefore we use 10 fold cv scheme to calculate each of these measures to minimize bias for the test data, and to improve precision in the scenario of induced variance due to the data-driven steps in model building and validation measure. 

\subsubsection{C-index}
Harrell’s concordance index or C-index~\citep{harrell1982evaluating} is perhaps the most popular measure of model's discriminative strength in the right-censored survival analysis literature. In order to compute C-index, we first need to define permissible cases and concordant pairs. To account for the censoring, the set $\beta$ of permissible cases consists of all possible pairs of individuals, $i$ and $j$ in the data, but with two exceptions: 1) the ones in which shorter survival time is censored, and 2) when $T_i=T_j$, but neither of $i$ and $j$ has the event (death). Now, for any randomly selected pair out of the permissible cases, a pair can be concordant or partially concordant depending on their values of ensemble hazard, event time, and censoring status. For example, for a pair with distinct ensemble hazard and event times, a concordance value to 1 is assigned if the predicted risk (in terms of ensemble CHF) is greater for the individual that experiences death first i.e., $Pr\big(\sum_{l=1}^{n}\hat{H}_e(t^*_l|\bm{x_i})>\sum_{l=1}^{n}\hat{H}_e(t^*_l|\bm{x_j})|T_j>T_i \big)$. Here $t^*_1,...,t^*_n$ denote all the unique event times in $\bm{\Phi}$. For each pair in $\beta$, the concordant pairs and their assigned concordance values can be given as: 

\begin{eqnarray*}
\mathcal{I}=
\begin{cases}
1, & \begin{cases}
(\hat{H}_e(t^*_l|\bm{x_i})>\hat{H}_e(t^*_l|\bm{x_j})|T_j>T_i)\\
(\hat{H}_e(t^*_l|\bm{x_i})>\hat{H}_e(t^*_l|\bm{x_j})|T_j=T_i) ~\& ~(\delta_i=1, \delta_j=0)\\
(\hat{H}_e(t^*_l|\bm{x_i})=\hat{H}_e(t^*_l|\bm{x_j})|T_j=T_i) ~\& ~(\delta_i=\delta_j=1)\\
\end{cases}\\
0.5, & \begin{cases}
(\hat{H}_e(t^*_l|\bm{x_i})=\hat{H}_e(t^*_l|\bm{x_j})|T_j \neq T_i)\\
(\hat{H}_e(t^*_l|\bm{x_i})\neq\hat{H}_e(t^*_l|\bm{x_j})|T_j=T_i) ~\& ~(\delta_i=\delta_j=1)\\
(\hat{H}_e(t^*_l|\bm{x_i})=\hat{H}_e(t^*_l|\bm{x_j})|T_j=T_i) ~\& ~(\delta_i=1, \delta_j=0)\\
(\hat{H}_e(t^*_l|\bm{x_i}) < \hat{H}_e(t^*_l|\bm{x_j})|T_j=T_i) ~\& ~(\delta_i=1, \delta_j=0)\\
\end{cases}\\
0, & \text{otherwise}
\end{cases}
\label{eq:eq6}
\end{eqnarray*}

Then the C-index can be expressed as the ratio of the sum of concordance values and the total number of permissible pairs as:
\begin{eqnarray*}
C=\dfrac{\sum_{i,j \in \beta}\mathcal{I}}{|\beta|}
\label{eq:eq1}
\end{eqnarray*}

Since $C$ represents the classification probability of the model, a higher value is desirable. A value of 50 is essentially no better than random guessing.

\subsubsection{Prediction error curves (PEC)}
We use PEC to capture a model's prediction of the survival probability for the test data at different time points. In the absence of censoring, PEC for an individual $i$ in the test data is an expectation of the squared difference between the true survival status and predicted survival probability of $i$ at time $t$ with biomarkers $\bm{x}_i$. However, censoring introduces bias in the population average of PEC. The introduction of inverse probability of censoring weight (IPCW) by~\citet{gerds2006consistent} provides a versatile measure to overcome this limitation by weighting the squared residuals using IPCW. Given the survival data $\bm{\Phi}_i=\left\{\left(\bm{x}_i, T_i, \delta_i\right) \right\}_{1\leq i\leq N}$, let the test dataset $D_M$ contain $M$ independent and identically distributed replicates of $\bm{\Phi}_i$, where $M < N$. With the observed status for subject $i$, $\tilde{\mathcal{Y}}_i(t)= \mathbbm{1}_{{T}_i>t}$ and its predicted survival status $\hat{S}(t|\bm{x}_i)$, the prediction error or Brier score at time $t$ is given as: 
\begin{eqnarray}
E(t,\hat{S})=\frac{1}{M}\sum_{i \in D_M}\hat{W}_i(t)\left\{\tilde{\mathcal{Y}}_i(t)-\hat{S}(t|\bm{x}_i)\right\}^2
\label{eq:eq2}
\end{eqnarray}
In Equation~\ref{eq:eq2}, the inverse probability of the censoring weights is estimated as~\citep{gerds2007efron}: 
\begin{eqnarray*}
\hat{W}_i(t)=\dfrac{(1-\tilde{\mathcal{Y}}_i(t))\delta_i}{\hat{G}({T}_i-|\bm{x_i})}+\dfrac{\tilde{\mathcal{Y}}_i(t)}{\hat{G}(t|\bm{x_i})}
\label{eq:eq3}
\end{eqnarray*}
where $\hat{G}(t|x) \approx P(C_i >t|{x_i}=x)$ denotes the estimated conditional survival function of the censoring time. Here, the prediction error in Equation~\ref{eq:eq2} is estimated from test data via a 10 fold cv scheme. The aim here is to give the averaged prediction error at every time point in the test data. We also use survival probability plots of individuals in the test data at all event time points to show the predicted survival probability of the balanced and unbalanced models.

\subsubsection{Integrated Brier Score (IBS)}
IBS consolidates the PEC estimates over all time points and is defined as:
\begin{eqnarray*}
IBS(E, \tau)=\frac{1}{\tau}\int_{0}^{\tau}E(\mu,\hat{S})du
\label{eq:eq5}
\end{eqnarray*}
Where $\tau$ is the total time span for which the prediction errors can be estimated. Since it is an average of the PEC, we also desire small error. A value of 0.25 means that irrespective of their risk status, the model predicted $50\%$ risk for all the individuals whereas, a value of $0$ indicates perfect prediction.
%	\item{MSE}\cmmt{Time dependency}
%	\item{Sensitivity and specificity}
	
\subsection{Class Balancing Scheme}\label{balancing}
The survival data, in general, has highly imbalanced classes. This extreme imbalance results in suboptimal performance of the survival models, whether it be CPH or RSF. Further, the size of the survival is another concern as the data obtained from the tertiary care hospitals providing acute care are often small, which further aggravates the issue~\citep{japkowicz2002class}. Several balancing methodologies have been proposed in the literature to address class imbalance and have been applied in the context of random forests~\citep{chen2004using}, albeit, with very limited consideration in survival analysis. In this work, we emphasize the importance of balancing the survival data in order to develop an accurate prediction model. 

\begin{figure}[!htb]\centering
	\includegraphics[width=.45\textwidth]{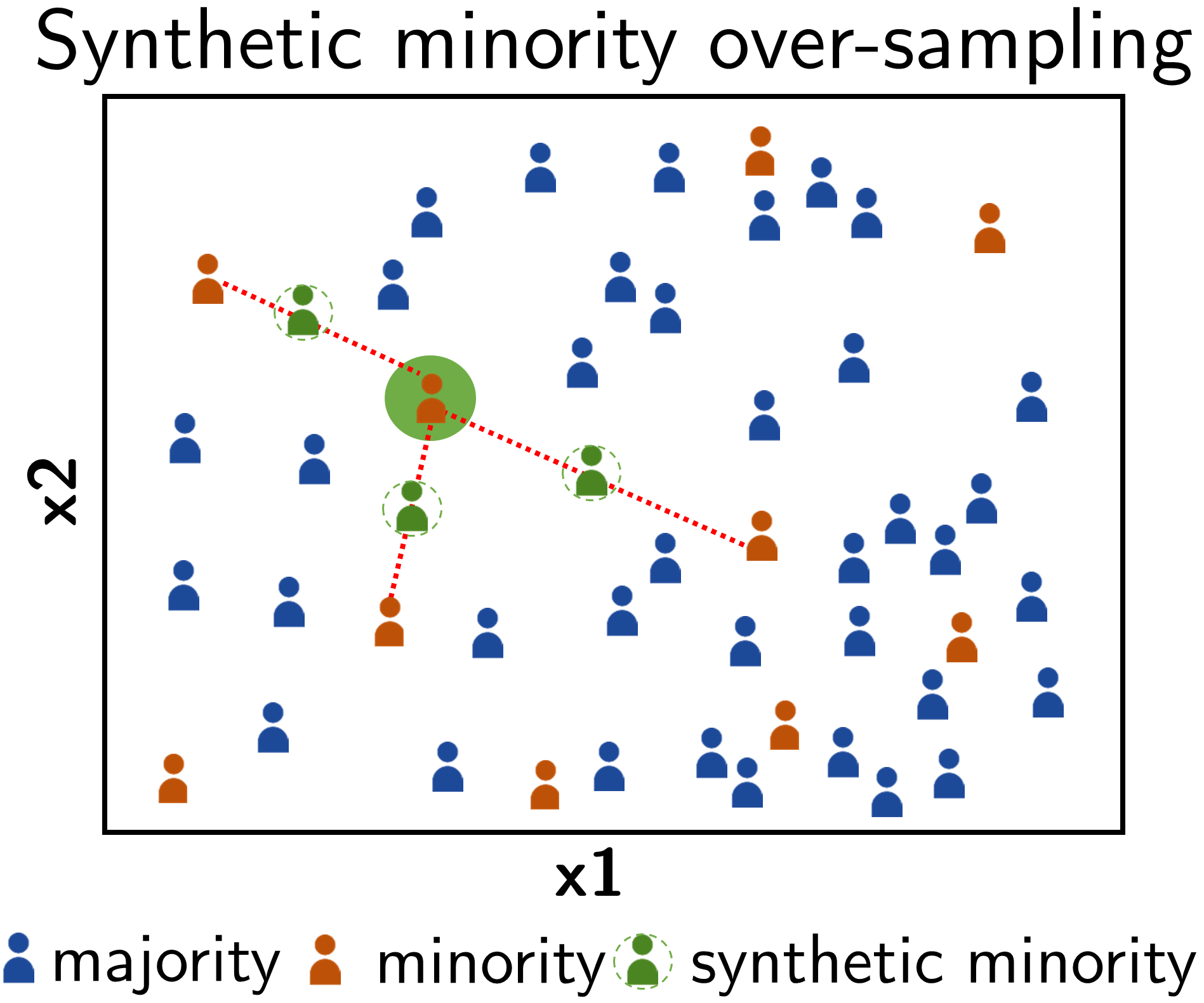}
	\caption{ Represents the imbalance in STEMI dataset and synthetically generated minority (in green) using SMOTE. }
	\label{fig:i1}
\end{figure}

Data balancing has been addressed using resampling methods such as under-sampling the majority class until their numbers are reduced or made equal to the number of samples in the minority class or over-sampling the minority class until its size is as large as the majority class. Prior investigations suggest that over-sampling does not improve the minority class representation significantly and under-sampling is a better approach than over-sampling~\citep{japkowicz2000class,chawla2002smote}. Unfortunately, various real-life scenarios, including the present context where data is obtained from tertiary care hospitals, only limited samples are available. In such cases, under-sampling leads to an unwanted decrease in the training dataset and is not a feasible option. The imbalance shown in Figure~\ref{fig:i1}(a) is representative of the STEMI dataset which consists of tracking 267 patients for their mortality over a period of 1 year, out of which 62 (only 23\%) belong to the minority class (i.e., suffered mortality). This led us to explore a synthetic generation of minority class samples without resorting to excessive under-sampling. We adopt the synthetic minority over-sampling technique (SMOTE) proposed by \citet{chawla2002smote}.
\begin{figure}[!htb]\centering
	\includegraphics[width=.8\textwidth]{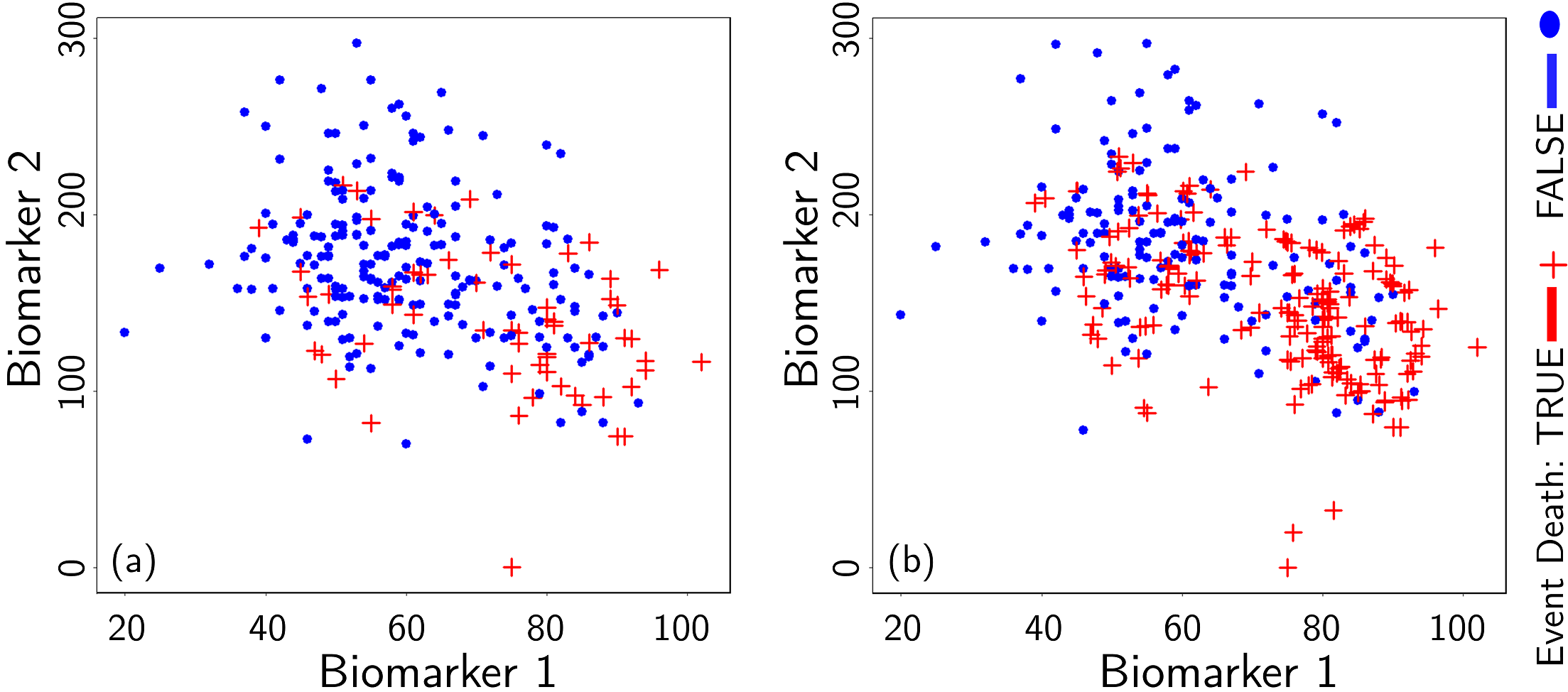}
	\caption{(a) Representation of the class imbalance in the biomarker space (minority class in red) (b) balanced class representation using synthetically generated minority  }
	\label{smote}
\end{figure}

This synthetic generation process proceeds by randomly selecting a minority and its $k$ nearest minority class neighbors. The value of $k$ is determined by the amount of over-sampling needed. Let $\bm{x}_i$ be the feature vector representing the biomarkers for the selected minority and $\bm{x}_j$ be the feature vector of a randomly chosen neighbor, then a new synthetic minority, $\bm{x}_s$ is generated in the biomarker (feature) space as follows: 
\begin{eqnarray}
	\bm{x}_s= \bm{x}_i+\varGamma\left(\bm{x}_i-\bm{x}_j\right)
\end{eqnarray}
where, $\varGamma \sim \text{Uniform}(0,1)$ is a uniform random variable. Thus, the synthetically generated data can be interpreted as a randomly sampled point
along the line segment between the two minority samples in the biomarker space. Depending on the extremity of imbalance, a sample can be created along with all the lines joining the selected minority sample and its $k$ neighbors. Representation of this scheme in two-dimensional feature space is shown in Figure~\ref{fig:i1}. 

The following results outline the impact of extreme imbalance on RSF, and the effect of balancing in improving survival prediction. Proving any result related to the hazard estimation requires establishing the form of true cumulative hazard function. As defined above, ${T^0_i}_{\left\lbrace 1\leq i\leq N \right\rbrace }$ are the actual survival times for the $N$ individuals. These survival times are assumed to be dependent with common continuous marginal distribution function $F(t)=P(T_i \leq t)$. The underlying true cumulative hazard function is then given by $H(t)=-\log(1-F(t))$ \citep{cai1998asymptotic}.  In the remainder of this section, we first prove that the hazard estimates $\hat{H}_M(t)$ and $\hat{H}_C(t)$ are underestimations of the true hazard whenever the minority class is mortality and an overestimation in case the survival/censored class is minority. Subsequently, we establish the theoretical results for improvement in the prediction error after balancing.

\begin{proposition}
Let $m_1$ and $m_2$ be the number of censored and mortality samples in $\bm{\Phi}$ and $\hat{H}_C(t)$, $\hat{H}_M(t)$ be the cumulative hazard function estimated for the censored and mortality nodes respectively. Then for the mortality node $\hat{H}_M(t)$ is an underestimation, i.e., $\hat{H}_M(t)<H(t)$ for $m_2 << m_1$ and overestimation $(\hat{H}_C(t)>H(t))$ for $m_1<< m_2$.
\label{prop1}
\end{proposition}

These results are based on two important aspects of the RSF construction, namely, 1) daughter node size constraint, and 2) terminal node hazard estimation (see Section~\ref{section 2}). The daughter node size constraint states that each of the resulting daughter nodes after the split must contain a minimum of $d_0$ unique deaths. The splitting terminates if the criterion is not satisfied. When the size of the mortality class is very small as compared to the censored class, the tree terminates prematurely and might result in decision region to be smaller and biased. As a default setting of the RSF, a minimum of $d_0>3$ unique deaths needs to be present in each of the daughter nodes. Since each case in a particular terminal node has the same hazard function as the cumulative hazard function of the constituting terminal node, the hazard for the death samples in the terminal node with censoring majority is lower. Since the mortality samples are originally smaller in number this results in underestimation of their overall hazard function. 

Balancing further results in an improvement in the prediction error. This improvement in the prediction error (in terms of Brier score) from $\rho(t)$ to $\rho'(t)$ after balancing can be stated using the following proposition which considers the present context where the number of the mortality (minority) class samples are much fewer that that of the survival (majority) class. Here we assume $m_2<<m_1$ and an almost perfect split (for simplicity of calculation): 
\begin{proposition}
	For $m_2<<m_1$, let $\left\{m_1,m_2,\rho(t)\right\} $ and $\left\{{m}_1,{m}'_2,{\rho}'(t)\right\}$ be the surviving and the mortality class size, and the Brier score (BS) before and after balancing, respectively, and $d_0$ be the minimum number of unique death (mortality class) samples needed to present in the leaf nodes of an RSF tree. Assuming an almost perfect split with $m_2-d_0$ samples in the mortality node and $m_1+d_0$ samples in the censoring node (details in~\href{app}{Appendix}), ${\rho}'(t)$ can be approximated as: 
	\begin{eqnarray*}
	{\rho}'(t)= \rho(t)\bigg(\dfrac{m_1+m_2}{{m}_1+{m}'_2}\bigg) \left\{\dfrac{({m}'_2-d_0)e^{-2\hat{H}'_M(t)}+d_0e^{-2{\hat{H}_C(t)}}+{m}_1(1-e^{-{\hat{H}_C(t)}})^2 }{(m_2-d_0)e^{-2\hat{H}_M(t)}+d_0 e^{-2\hat{H}_C(t)}+{m}_1(1-e^{-\hat{H}_C(t)})^2 }\right\}
	\end{eqnarray*}
\label{prop2}
\end{proposition} 

The result can be similarly derived for the case when $m_1<<m_2$. This proposition leads us to our next result on how the unbalanced Brier score or prediction error related to the balanced error.\\
\noindent{\bf Corollary 1}{
	Let $\left\lbrace \rho(t), {\rho}'(t)\right\rbrace$  be the Brier scores before and after balancing the class sizes, then ${\rho}'(t)<\rho(t)$.
	\label{corollary1}
}
This Corollary establishes that after addressing the class imbalance, the prediction error decreases. Proofs for the Proposition~\ref{prop1}~\&~\ref{prop2}, and Corollary~\ref{corollary1} are provided in the~\href{app}{Appendix} of this paper. \\

\section{Case Studies } \label{case}
We apply BRSF on six different real-world survival datasets with varying degree of class imbalance. The use of real-data for BRSF's comparative analysis is to ascertain its relative effectiveness and suitability in the real world decision makings. Our main point of focus in this study is the STEMI dataset obtained from Heart, Artery, and Vein Center of Fresno. This dataset

\subsection{Performance Evaluation on Benchmarking Datasets}
Five of the six datasets (except the STEMI dataset) used in this study were obtained from online repositories, each with a different level of imbalance. These 5 datasets consists of survival analysis data for acute diseases such as lung cancer (veteran and lung datasets), a rare and fatal chronic liver disease (pbc dataset), acute stroke in patients with atrial fibrillation (COST dataset), and plasma cell immune disorder which may result in malignancy (mgus dataset). A summary of the class proportions in all the datasets for the censored and the event classes is given in Table~\ref{table:t1}. 

	\begin{table}[!htb]\centering
	\renewcommand{\arraystretch}{1.3}
	\caption{Summary of the real-world data sets used for model evaluation}
	\begin{threeparttable}
	\begin{tabular}{l  c  c c  }
		\toprule
		& &\textbf{ Class proportions} & \\ \cmidrule{2-4}
		\text{\textbf{Dataset}} & \text{Total} & \text{Censored} & \text{Event}  \\
		\hline 
		\text{veteran} \citep{kalbfleisch2011statistical} & 137 & {\color{red}9} & 128 \\ 
		\text{mgus} \citep{kyle1993benign} & 241 & {\color{red}16} & 225 \\
				\text{COST} \cite{jorgensen1996acute}& 518 & {\color{red}114} & 404\\
						\text{STEMI} \citep{sawant2013prognostic}& 267 & 205 & {\color{red}62}\\
								\text{lung} \citep{loprinzi1994prospective} & 228 & {\color{red}63} & 165\\
										\text{pbc} \citep{ishwaran2007random} & 418& 257 & {\color{red}161}\\
		\bottomrule
	\end{tabular}
\begin{tablenotes}\footnotesize
\item[*] Minority class represented in red
\end{tablenotes}
	\end{threeparttable}
	\label{table:t1}
\end{table}

Most of the datasets contained several missing values which were then imputed using adaptive tree imputation~\citep{ishwaran2008random}. To compare CPH and RSF and to determine the effect of balancing on these models, we use the C-index and IBS measures described in subsection~\ref{error}. Table~\ref{errorall} presents the average C-index and IBS scores for CPH, balanced CPH (BCPH), RSF, and BRSF obtained via a 10 fold cv scheme. The best model obtained for both C-index and IBS are shown in blue. As evident from this Table, BCPH and BRSF consistently perform better than their unbalanced counterparts. Additionally, the performance of BRSF supersedes all other models. 

\begin{table}[!htb]\centering
	\renewcommand{\arraystretch}{1.3}
	\caption{Performance evaluation results for the benchmark datasets}
	\begin{threeparttable}
	\begin{tabular}{l  c  c c c c }
		\toprule
		& &\textbf{ Model} & &\\ \cmidrule{2-6}
	\text{\textbf{Dataset}} &	\text{\textbf{Error measure}} & \text{CPH} & \text{BCPH} & \text{RSF}  & \text{BRSF}   \\
		\hline 
%veteran data 
	\text{\textbf{veteran}}&	\text{C-index} & 59 (0.24)  & 58 (0.21)& 61 (0.11) & {\color{blue}{77 (0.04)}} \\ 
	&	\text{IBS} & 0.15 (0.04) & 0.14 (0.03)& 0.15 (0.05) &{\color{blue} {0.09 (0.02)}}\\
\hline
% mgus data 
	\text{\textbf{mgus}}&	\text{C-index} & 71 (0.05)  & {\color{blue}{89 (0.021)}}& 69 (0.07) & 88 (0.02) \\ 
	&	\text{IBS} & 0.13 (0.02) & 0.06 (0.01)& 0.14 (0.02) &{\color{blue} {0.04 (0.01)}}\\
\hline
% Cost data 
	\text{\textbf{COST}}&	\text{C-index} & 69 (0.03)  & 76 (0.03)& 64 (0.04) & {\color{blue}{85 (0.01)}} \\ 
	&	\text{IBS} & 0.17 (0.01) & 0.15 (0.01)& 0.18 (0.02) &{\color{blue} {0.06 (0.01)}}\\
\hline
% lung data 
	\text{\textbf{lung}}&	\text{C-index} & 61 (0.09)  & 70 (0.04)& 59 (0.09) & {\color{blue}{76 (0.03)}} \\ 
	&	\text{IBS} & 0.18 (0.01) & 0.13 (0.01)& 0.18 (0.02) &{\color{blue} {0.08 (0.01)}}\\
\hline
% pbc data 
	\text{\textbf{pbc}}&	\text{C-index} & 77 (0.09)  & 79 (0.02)& 78 (0.08) & {\color{blue}{83 (0.02)}} \\ 
	&	\text{IBS} & 0.14 (0.02) & 0.12 (0.01)& 0.13 (0.02) &{\color{blue} {0.07 (0.01)}}\\
		\bottomrule
	\end{tabular}
\begin{tablenotes}\footnotesize
\item[*] Numbers inside the bracket represents standard deviation across 10 fold cv
\end{tablenotes}
	\end{threeparttable}
	\label{errorall}
\end{table}

Given this result, we now focus on the STEMI dataset obtained for the Heart, Artery, and Vein Center of Fresno to do the further in-detail analysis. A concise description of the study design and biomarkers for this data is provided in subsection~\ref{stemi} and the results of these analyses are then discussed in the subsequent sections. 
	
\subsection{STEMI Dataset Study Design and Biomarkers}\label{stemi}
The study cohort for the STEMI dataset consisted of 278 consecutive patients. The patients had electrocardiographic criteria for STEMI and a presumed diagnosis for acute coronary syndrome at the time of presentation to the emergency room of \textbf{ a tertiary care hospital in central California, USA}. Electrocardiographic, radiographic, and basic laboratory investigations were obtained at the time of presentation and an emergent coronary angiography was performed. Patients underwent coronary artery bypass grafting (CABG) or primary percutaneous coronary intervention. Enrollment into the study began in January 2007 and patient were followed for one year until January 2008. A detailed design of this retrospective study has previously been published \citep{sawant2013prognostic}. We focused primarily on $N=267$ patients (187 male and 80 female) who did not have preexisting left bundle branch block or paced rhythm on ECG. Dataset consisted of a large set ($R=150$) of biomarkers. These biomarkers included therapy provided, physiological and anatomical variables such as age, gender, ethnicity, BMI, ECG criteria, the occurrence of cardiac arrest during admission, troponin levels at the time of discharge, Brain Natriuretic Peptide (BNP) levels, and clinical risk measures such as TIMI index, Mayo Clinic risk score etc. along with the previously mentioned laboratory measurement. The dataset had ethnically diverse population including Black, Caucasian, and a high percentage of representative minority populations such as American Indian, Asian and Hispanics thus somewhat offsetting the demerits of small data size. Mortality data were obtained either from either the hospital, California Department of Public Health (CDPH) or Social Security Death Index records. To avoid any confounding effects of loss to follow-up, and accurate determination of the cause of the death, an all-cause mortality was selected as a primary endpoint instead of disease-specific mortality. Out of the 267 patients,62 patients died in one-year duration (representing the minority class for this dataset).

\subsection{Performance Evaluation on STEMI Dataset} \label{section 4}
\begin{figure}[!htb]
	\includegraphics[width=1\textwidth]{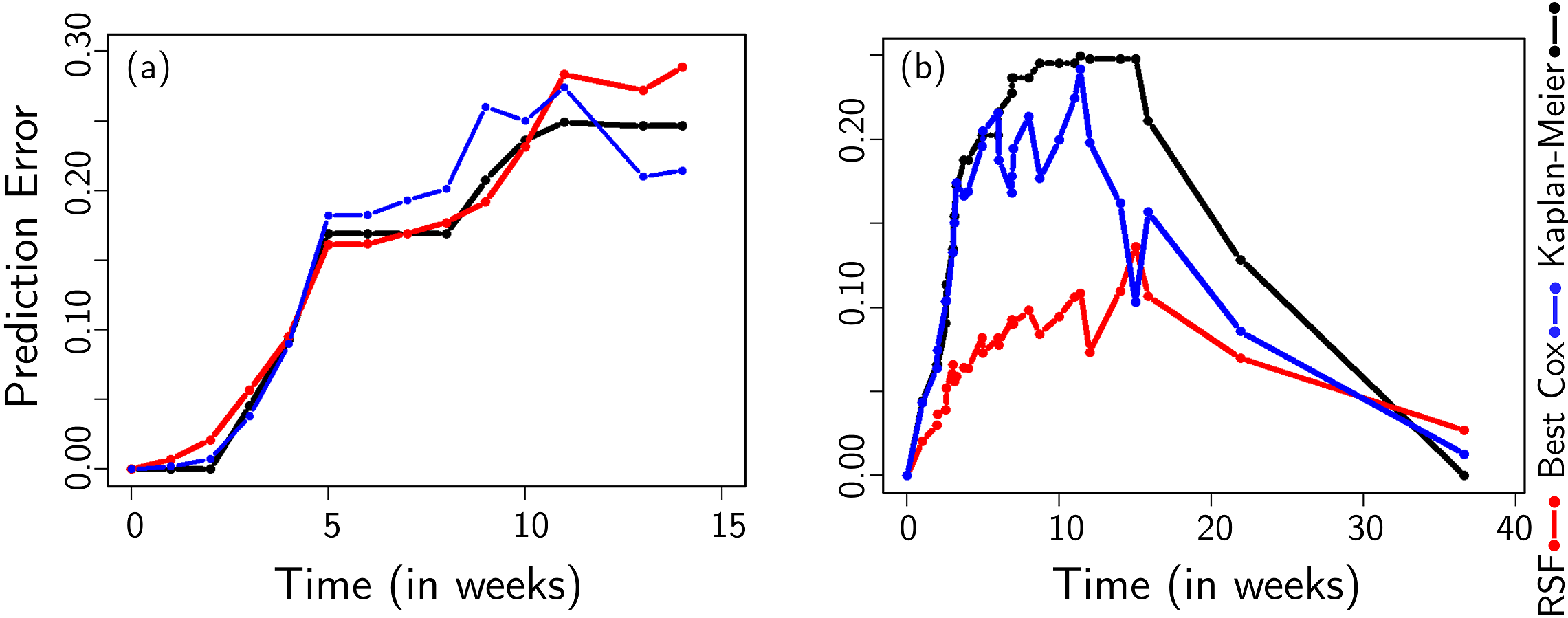}
	\caption{Prediction Error Curves (PEC) for RSF (blue), Best Cox (red), and reference Kaplan-Meir (black) (a) unbalanced (b) balanced }
	\label{pec}
\end{figure}

We again evaluate the CPH, BCPH, RSF, and BRSF models with respect to C-index and IBS scores obtained via 10 fold cv on the STEMI dataset. As can be seen from the Table~\ref{table:t2}, presenting the average performance of the models,the balanced models perform better than their unbalanced counterparts with an exception of balanced CPH which is statistically the same as unbalanced CPH in terms of C-index. In terms of IBS, BRSF performs significantly better with a $53\%$ improvement than the unbalanced RSF. In Figure~\ref{pec}, we show this improvement in IBS score by plotting prediction error or Brier score at various times points of the 1 year study duration for one of the 10 folds of cv trials. 
	\begin{table}[!htb]\centering
	\renewcommand{\arraystretch}{1.3}
	\caption{Predictive performance evaluation results}
	\begin{tabular}{l  c  c c c }
		\toprule
		& &\textbf{ Model} & &\\ \cmidrule{2-5}
		\text{\textbf{Error measure}} & \text{CPH} & \text{BCPH} & \text{RSF}  & \text{BRSF}   \\
		\hline 
		\text{C-index} & 80 (0.12)  & 79 (0.05)& 82 (0.08) & {\color{blue}{82 (0.06)}} \\ 
		\text{IBS} & 0.18 (0.07) & 0.12 (0.04)& 0.17 (0.06) &{\color{blue} {0.08 (0.01)}}\\
		\bottomrule
	\end{tabular}
	\label{table:t2}
\end{table}

\begin{figure}[!htb]
	\includegraphics[width=1\textwidth]{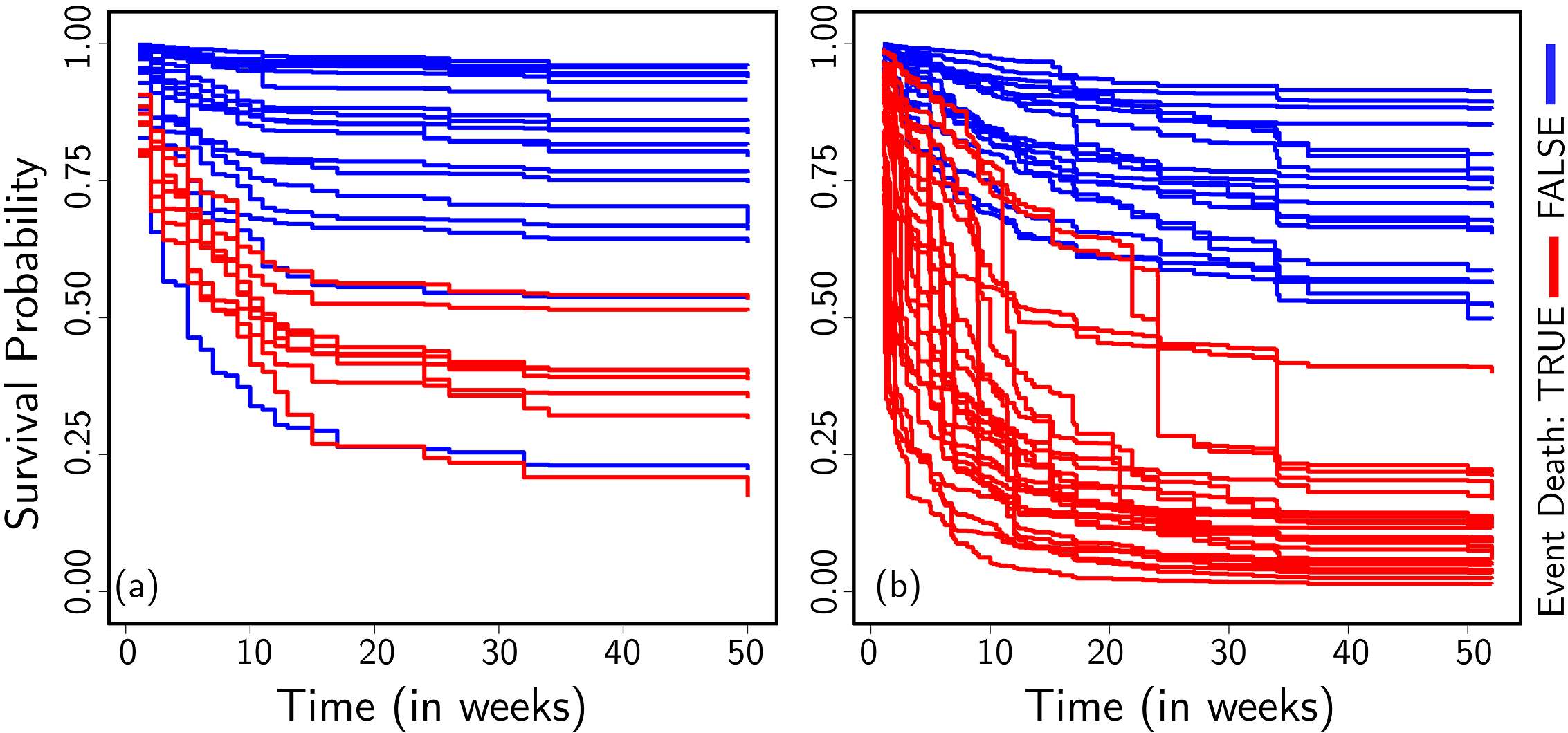}
	\caption{Survival Probability Curves for death (in red) and surviving samples (in blue) (a)RSF (b) BRSF}
	\label{surv}
\end{figure}

\begin{figure}[!htb]
	\includegraphics[width=\textwidth]{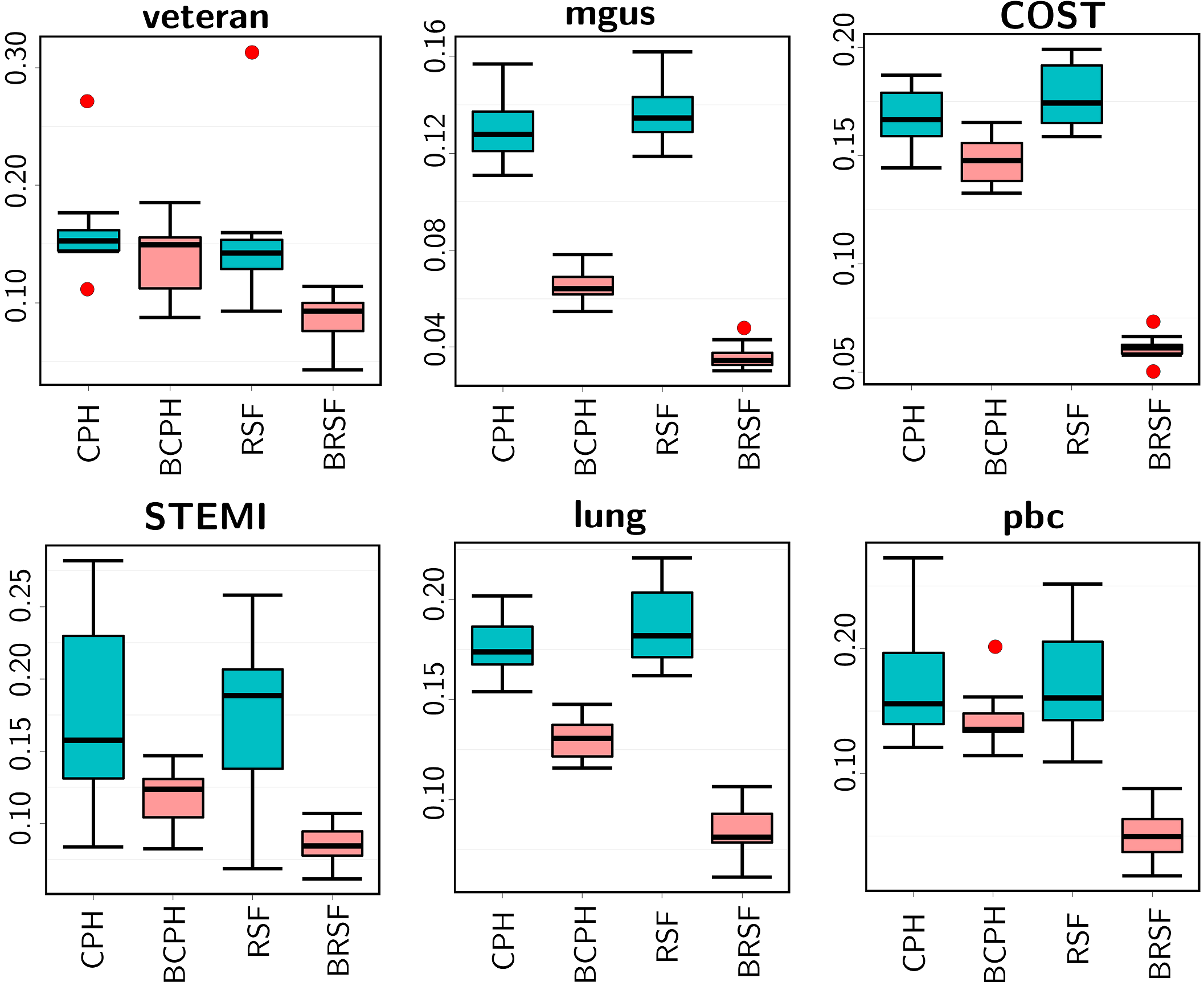}
	\caption{Boxplots of estimated IBS calculated for the test data in 10-fold CV scheme for $6$ data sets arranged in decreasing order of class imbalance. The horizontal line inside the box represents the median and the box is bounded by the $25^{th}$ and $75^{th}$ percentile (IQR). whiskers extend to $1.5\times\text{IQR} $ and the outliers are represented by the red dot.}
	\label{pecall}
\end{figure}

Further, this performance can be better represented in terms of the survival curves. The survival probability for the STEMI test samples in the unbalanced and balanced data are shown in Figure~\ref{surv} (a) and (b) respectively. When the classes are balanced not only their separability (i.e. higher survival probability for the false event and lower for the true event) increases the survival/hazard estimates for the minority samples also improves. The survival probability plots for all other datasets (Table~\ref{table:t1}) are shown in the Appendix~\ref{app} demonstrating similar improvement in the survival probability estimates. Additionally, Figure~\ref{pecall} summarizes the 10 fold cv IBS for all the datasets. There was an overall improvement of $25\%$ in the C-index and $55\%$ in the IBS score from RSF to BRSF. 

We obtained 7 best biomarkers based on backward selection. Their OOB error was 15.8\% compared to 18\% with all 150 predictors. It turns out these predictors have high importance per both Breiman's variable importance (VIMP)~\citep{breiman1984classification} and Ishwaran et. al.'s minimal depth (MD) scores~\citep{ishwaran2010high} (see Table~\ref{variables}). From a physiological standpoint, these covariates are among the most significant biomarkers of survival during acute cardiac diseases, as elaborated in the following paragraphs.

	\begin{table}[!htb]\centering
		\renewcommand{\arraystretch}{1.3}
		\caption{Biomarker Selection: Top 7 Biomarkers}
		\begin{tabularx}{\textwidth}{l  c c c c c c c  }
			\toprule
			& & & &\textbf{Biomarkers}  & & &\\ 
			\cmidrule{2-8}
			\text{\textbf{Ranks}} & \text{Disch}& \text{MCRS}&\text{Cron}& \text{GRACE} & \text{MCRS} & \text{CHF}& \text{ACS}        \\
			\text{\textbf{(statistics)}}& \text{Trop} & \text{MS}&\text{DC}& \text{Prob}&\text{MACE} &  \text{ in1yr}& \text{in1yr}       \\
			\hline 
			\text{MD} 	&1(8.16) & 2(8.39) & 3(8.53)  &4(8.75)  & 7(9.28) &  8(9.30)  & 12(9.69)    \\
			\text{VIMP} &9(0.01) & 3(0.02) & 1(0.02)  & 8(0.01)  & 7(0.01) & 4(0.02)  & 5(0.02)    \\ 
			\bottomrule
		\end{tabularx}
		\label{variables}
	\end{table}

\begin{figure}[!htb]
		\includegraphics[width=1\textwidth]{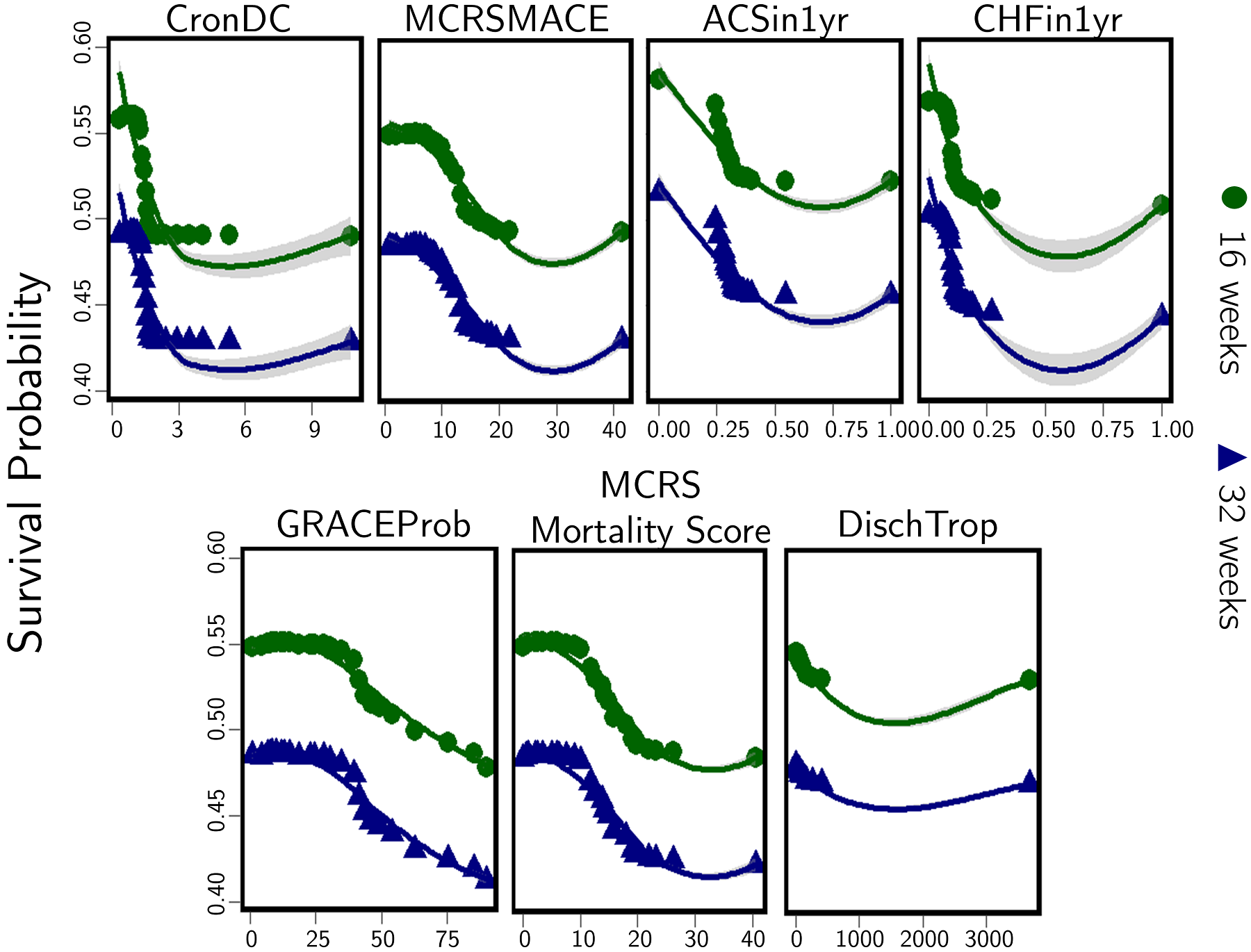}
		\caption{Partial dependence plot of predicted survival probability plotted as a function of top $6$ biomarkers for randomly chosen $100$ subjects (market as ``circle" and ``triangle"). The green and blue lines shows the trend at $16^{th}$ and $32^{nd}$ weeks respectively }
		\label{partialplot}
	\end{figure}

We graphically explored the relation of thus selected most important biomarkers (top 7) with the survival probability using partial dependence plots and verified their physical meaning and significance. In a survival setting, a partial dependence plot represents the response corresponding to the biomarker of interest at a particular time by averaging out the joint effect of remaining biomarkers~\citep{friedman2001elements,ehrlinger2016ggrandomforests}. In Figure~\ref{partialplot}, the two curves corresponding to each of the biomarkers shows the trend of survival probability with changing value of the biomarker at $16^{th}$ and $32^{nd}$ week for $100$ randomly chosen subjects. It shows nonlinearly-decreasing survival probability with increasing value of ``CronDC", ``MCRS-Mortality Score(MS)",``MCRS-MACE", ``CHFin1yr",``GRACEProb", ``DischTrop" (NOTE: Since all plots have same vertical axis limits, some dominant non-linear relationships make curves for ``GRACEProb" and ``MCRS-MS" appear flatter than they actually are). For all the biomarkers, we can see the decreasing survival probability with increasing time (``blue" line for $32^{nd}$ week is below the ``green" line for $16^{th}$ week). The variables selected were evaluated by a cardiologist to have a significant physical correlation with the prediction of mortality.

\begin{figure}[!htb]\centering
		\includegraphics[width=.9\textwidth]{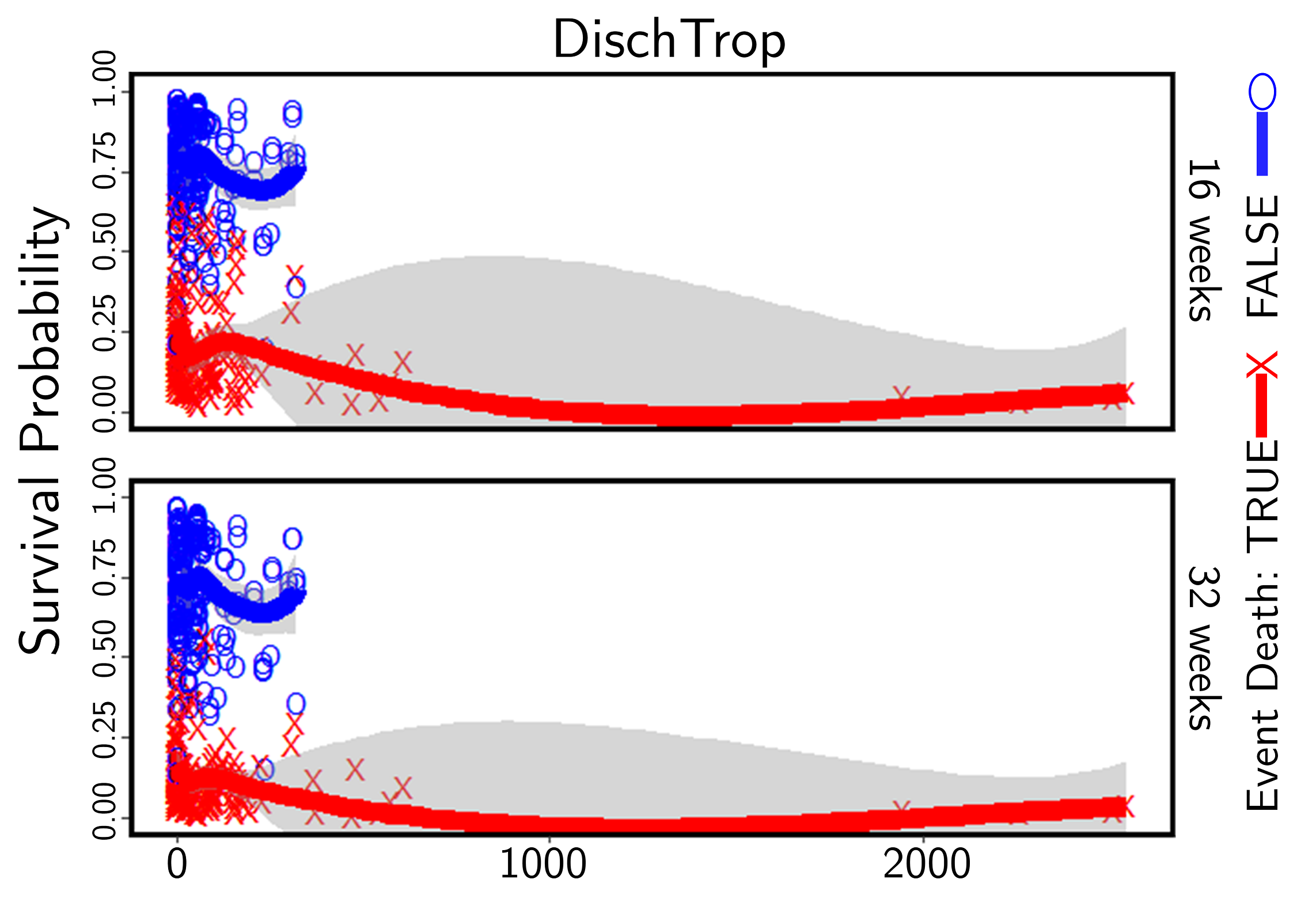}
		\caption{Variable dependence plot of survival probability plotted as a function of DischTrop at $16^{th}$ and $32^{nd}$ weeks}
		\label{variableplot}
	\end{figure}

In particular, the DischTrop (discharge troponin) biomarker, recording the troponin levels during the patient discharge is studied as a primary diagnostic component~\citep{ottani2000elevated, croal2006relationship}. Troponin is a protein released during myocardial infarction. A higher level of troponin indicates more damage to the cardiac muscle. Figure~\ref{variableplot} is a visualization of survival probability trend with varying DischTrop level across the mortality and survival samples. Patients with higher level of troponin content during the discharge shows to have lower survival probabilities. Though application of machine learning algorithms, in particular, RF based approaches are often criticized for their lack of interpretability in real-world data, the variable and partial dependence plots for all the biomarkers can provide insightful information on their relationship with mortality. Consequently, the proposed technique can be used can be used by a healthcare practitioner as an analytical analysis tool to achieve improved throughput and accuracy.

\section{Conclusions} \label{summary}
In this paper, we have introduced a BRSF model for survival analysis which is expected to address the limitations of both traditional methods such as CPH and newer methods such as RSF in handling extremely imbalanced datasets. The theoretical results as well as extensive experimental analysis presented in this paper demonstrate the superiority of BRSF method in terms of various performance measures. Empirical studies with $6$ datasets suggest a $55\%$ improvement in IBS score. Although class imbalance has been extensively studied in the machine learning literature, its theoretical analysis and application in the domain of survival analysis still remain largely unexplored. Pertinently, this is among the first investigations into the effect of class imbalance on the performance of survival models. Specifically, the theoretical results on performance improvement accrued from balancing the RSF models as well as the detailed empirical studies can lead to further improvements to the algorithms for RSF as well as more optimized balancing strategies. This study may provide a foundation for further knowledge discovery and subsequent improvement in survival analysis---a healthcare domain of immense importance. 

\acks{We sincerely thank the Heart, Artery, and Vein Center of Fresno, CA for their effort in data collection used in this study. One or two? of the author's were supported by NSF-PFI-AIR-TT 1543226 and NSF CMMI 1301439 during this research.}

\newpage

\appendix
\section*{Appendix A.~Investigation of the Effect of Class Balancing on Survival Analysis}\label{app}

In this appendix we provide the proofs for Proposition 1, Proposition 2, and Corollary 1 \textbf{introduced in subsection~\ref{balancing} of the main text}. Additionally, supporting empirical results on the survival probability plots of the benchmark datasets obtained using RSF and BSRF are provided. 

\noindent
{\bf Proposition 1}{Extreme imbalance between the censored and mortality classes leads to underestimation (overestimation) of the cumulative hazard for the mortality (censored) class terminal nodes.}\label{lemma_1}

\noindent
{\bf Proof}. Growing a survival tree of RSF proceeds with recursively splitting of the tree nodes into daughter nodes such that the survival difference between the daughter nodes is maximized. In doing so, the ultimate goal is to grow a survival tree (and thus the forest) where each node is populated with homogeneous survival population. The caveat here is that each of the daughter nodes must contain $d_0>0$ unique deaths. Not fulfilling this criterion leads to the termination of the tree growth. At this point, there are $\mathcal{L}(\mathcal{T}_b)$ terminal/leaf nodes in the tree, $\mathcal{T}_b$. Let, $t_{1,h} <t_{2,h}<...<t_{N(h),h}$ be $N(h)$ ordered, unique event (death) times in the terminal node, $h \in \mathcal{L}(\mathcal{T}_b)$, then the CHF for individuals in this node is given using the Nelson-Aalen estimator as:  
\begin{eqnarray}
\hat{H}(t|\bm{x}_i)=\hat{H}_h(t)=\sum_{t_{l,h}\leq t}\dfrac{d_{l,h}}{Y_{l,h}},\quad \text{if}~\bm{x}_i \in h
\label{hazard}
\end{eqnarray}
In Equation~\ref{hazard}, $d_{l,h}$ and $Y_{l,h}$ represent, respectively, the number of deaths and the number of patient at risk in node $h$ at times $\left\{t_{l,h}\right\}_{1\leq l \leq N(h)}$. Since the construction survival tree is based on binary splits, $\bm{x}_i$ corresponding to each individual $i$ ends up in a unique leaf node of $\mathcal{L}(\mathcal{T}_b)$. Forest ensemble hazard for the individual is an average across all such leaf nodes in the forest. Further, in practice the trees are grown using bootstrap data which needs to be considered while estimaing the ensemble hazard (for details of growing RSF and estimating ensemble hazard please see section 2.1 of the main text). Nonetheless, the key point here is that determining the ensemble hazard and proving related result reduces to demonstrating them for a single leaf node. For further simplicity and easy interpretability of the steps shown in the proof, we define a possible best random survival split and employ it in our calculations. Let the parent node has $m_1$ deaths and $m_2$ censored samples. Then the terminal nodes of the best split have following conditions: (i) the survival nodes has exactly $d_0$ deaths (R package implementation of RSF has $d_0=3$) and (ii) except for the $d_0$ death samples, both the nodes have a homogeneous population. Thus, the survival leaf node has $m_1+d_0$ samples and the death leaf node has $m_2-d_0$ samples. Note that, with this construction, $m_2\geq 2d_0$. Further, both leaf nodes have distinct event times. Although we use a simple node split, the results can be adapted to a generalized tree construction and hence to the RSF. 

\begin{figure}[!htbp]\centering
	\includegraphics[width =.35\textwidth]{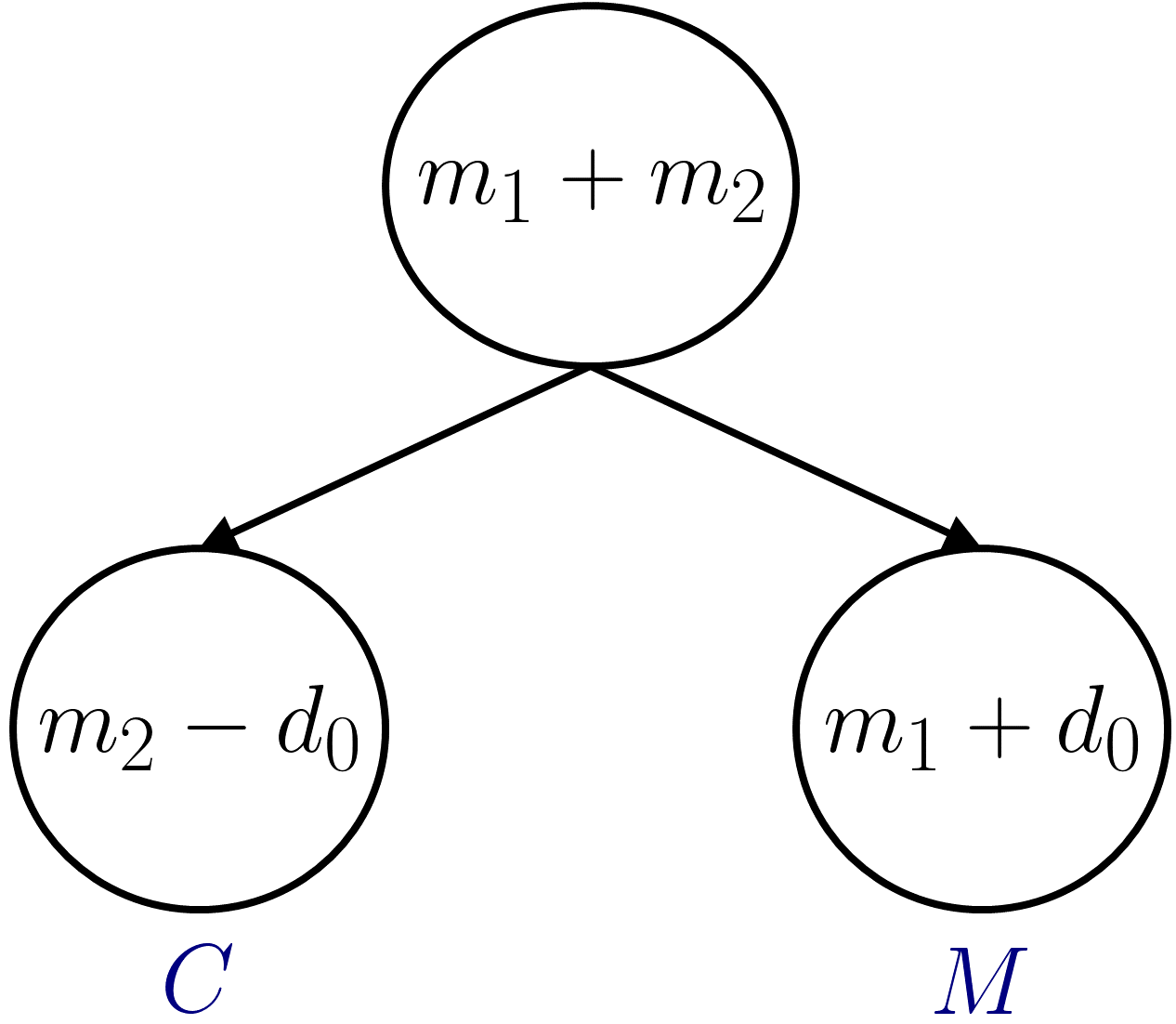}
	\caption{ A possible representation of best binary survival tree node split}
	\label{node}
\end{figure}

Let $M$ and $C$ denote the mortality and censoring/survival nodes respectively. For ease of calculation, we estimate the cumulative hazard of the nodes at their respective maximum event times. Let $t^{*}$ be the maximum event time at node $M$ then $\hat{H}_{M}(t^{*})$ is given as: 
\begin{eqnarray*}
	\hat{H}_{M}(t^{*})&=& \sum_{t_{l,{M}}\leq t^{*}}\dfrac{d_{l,{M}}}{Y_{l,{M}}}\\
	&=& \dfrac{1}{(m_2 - d_0)-1}+\dfrac{2}{(m_2 -d_0)-2}+...+ \dfrac{(m_2-d_0-1)}{1}
\end{eqnarray*}
Let $m_2-d_0=y$, then $\hat{H}_{M}(t^{*})$ can be represented as:
\begin{eqnarray}\label{deathnodehazard}
\hat{H}_{M}(t^{*}) &=& \bigg(\dfrac{1}{y-1}+ \dfrac{2}{y-2}+ ... + \dfrac{y-1}{1}\bigg)
\end{eqnarray}
Equation~\ref{deathnodehazard} can alternately be written as harmonic series as follows: 
\begin{align*}
	z^{y-1}_1&= \dfrac{1}{y-1}+\dfrac{1}{y-2}+\dfrac{1}{y-3}\dots+\dfrac{1}{2}+\dfrac{1}{1}&&\\
	z^{y-2}_2&=\dfrac{1}{y-2}+\dfrac{1}{y-3}+\dots +\dfrac{1}{2}+\dfrac{1}{1}&&\\
	z^{y-3}_3&= \dfrac{1}{y-3}+\dfrac{1}{y-4}+\dots+\dfrac{1}{2}+\dfrac{1}{1}&&\\
	&\vdotswithin{=}\\
	z^2_{y-2}&= \dfrac{1}{2}+\dfrac{1}{1}&&\\
	z^1_{y-1}&= \dfrac{1}{1}&&\\
\end{align*}

The sum of the $1^{st}$ series, $z_1^{y-1}$ with $(y-1)$ elements can be approximated using the following: 
\begin{eqnarray*}
z_1^{y-1}= \sum_{n=1}^{y-1}\dfrac{1}{n}=\gamma+\psi_0((y-1)+1)=\gamma+\psi_0(y) 
\end{eqnarray*}
Where, $\gamma\approx0.577$ is the Euler-Mascheroni constant \citep{lagarias2013euler} and $\psi_0(\mathord{\cdot})$ is the diagmma function \citep{abramowitz1972handbook}. Similarly,~$z_2^{y-2}=\gamma+\psi_0(y-1)$ and so forth. Hence, the hazard estimate for the mortality node can be given as: 
\begin{eqnarray}
	\hat{H}_{M}(t^{*})&=&(y-1)\gamma+\sum_{i=2}^{y}\psi_0(i) \nonumber\\
&=&(m_2-d_0-1)\gamma+\sum_{i=2}^{m_2-d_0}\psi_0(i)
\label{hazardmortality}
\end{eqnarray}

Similarly the hazard function for the censoring node, $C$ estimated at or after the maximum event time, $t^{**}$ considering the  minimum censoring time to be greater than the maximum event time, $\hat{H}_{C}(t^{**})$ can be represented as:
\begin{eqnarray*}
 \hat{H}_{C}(t^{**})= \dfrac{1}{m_1/2 + (d_0-1)}+\dfrac{2}{m_1/2 + (d_0-2)}+...+ \dfrac{d_0}{m_1/2}
 \label{censoredhazard}
\end{eqnarray*} 

It can be shown that the hazard estimate or survival function of the terminal nodes and thus the RSF is consistent~\citep{ishwaran2010consistency}. As already defined, $H(t)=-\log(1-F(t))$ denote the true cumulative hazard function with $F(t)$ being the density estimate of true survival times ${T^0_i}$, i.e., $F(t)=P(T^0_i \leq t)$. Then for a possible infinite time $\tau$ such that the hazard estimate at $\tau$ is finite, using the consistency of Kaplan-Meir estimator, we have, $$\underset{0 \leq t \leq \tau}{\sup} |\hat{H}_M(t)-H(t)| \overset{p}\to 0 \quad \text{as} \quad m_2 \to \infty$$ with convergence rate $\log(\log(m_2))/m_2$. However, let us consider a class imbalance with $m_2<<m_1$. Then, $m_2-d_0$ mortality samples have hazard $\hat{H}_M(t^{*})$ and the remaining $d_0$ samples have a small hazard of the censored node $\hat{H}_C(t^{**})$. However, $\hat{H}_C(t^{**})<\hat{H}_M(t^{*})$ under any reasonable split (i.e. any censored node has more censored samples than mortality samples and vice versa for the hazard node). The overall estimate of the hazard for $m_2$ mortality samples is $(m_2-d_0)\hat{H}_M(t^{*})+d_0\hat{H}_C(t^{**})<m_2\hat{H}_M(t^{*})$  and is thus underestimated.
\hfill\BlackBox\\

\begin{sidewaysfigure}[!htbp]\centering
	\includegraphics[width=1\textwidth]{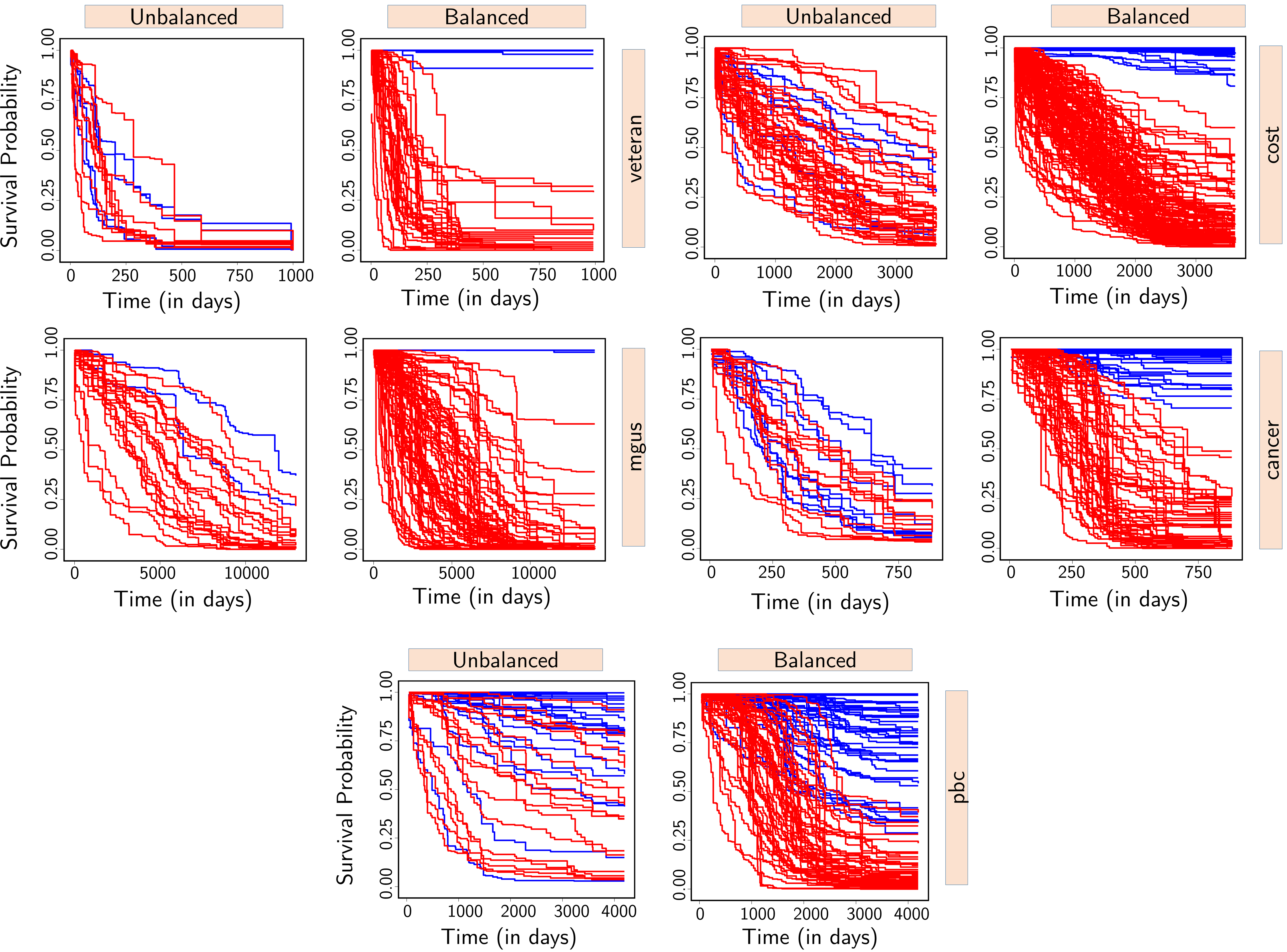}
	\caption{Survival probability plots for veteran, mgus, cost, cancer, and pbc data sets. For the pair of plots for each data set, the left plot represent the survival probability for original imbalanced data and the right one represent the survival probability after balancing with each class represented equally. }
	\label{Survival}
\end{sidewaysfigure}

Now, with additional synthetic mortality samples and the new mortality class size ${m}'_2 ({m}'_2> m_2)$, the hazard of the mortality class at $t^*$ becomes: 
\begin{eqnarray*}
	\hat{H}'_{M}(t^{*})=({m}'_2-d_0-1)\gamma+\sum_{i=2}^{{m}'_2-d_0}\psi_0(i)
\end{eqnarray*}
Also, $\hat{H}'_M(t^{*})-H_M(t^{*})= ({m}'_2- m_2)\gamma+(\sum_{i=2}^{{m}'_2-d_0}\psi_0(i)-\sum_{i=2}^{m_2-d_0}\psi_0(i))$. Clearly, hazard estimate of the individuals in the mortality node has now improved. Further, the $d_0$ mortality samples present in censored node still have hazard $\hat{H}_C(t^{**})$. Nonetheless, the proportion, $d_0/m_2>d_0/{m}'_2$, thus overall hazard of the individuals in the unbalanced, small size mortality class is underestimated. The underestimation is worsened when the size $m_2$ itself is small. Similarly, when $m_1<<m_2$ with an additional $d_0$ death samples in the censoring node, the unbalanced hazard $\hat{H}_C(t^{**})$ is overestimated. The hazard estimate of the censored node with ${m}'_1({m}'_1> m_1)$ can now be represented as: 
\begin{eqnarray}
\hat{H}'_C(t^{**})&=& \dfrac{1}{{m}'_1+d_0-1}+ \dfrac{2}{{m}'_1+d_0-2}+ ... + \dfrac{d_0}{{m}'_1}
\end{eqnarray}
Since ${m}'_1>m_1$, $\hat{H}_C(t^{**})>\hat{H}'_C(t^{**})$, i.e., the hazard for the censored node improves after balancing.
Figure~\ref{Survival}, presents the survival probability plots for the veteran, mgus, cost, and lung datasets for which censored class is the minority (pbc dataset has mortality as a minority, refer to Table~\ref{table:t1} in the main text). In this figure, the ``red curve" represents survival probability of the mortality samples and the ``blue curve" represents the survival of the censored samples at different event times. Ideally, the survival probability (approximately opposite of hazard) of the mortality should be low and that for the censored samples should be high. However, due to imbalance, there are several samples which are misclassified. After balancing, the blue curve has shifted upwards (hazard decreased). Further class separation has drastically improved. \\

%%%%%%%%%%%%%%%%%%%%%%%%%%%%%%%%%%%%%%%%%%%%%%%%%%%%%%%%%%%%%%%%%%%%%%%%%%%%%%%%%%%%%%%%
%% Proposition
%%%%%%%%%%%%%%%%%%%%%%%%%%%%%%%%%%%%%%%%%%%%%%%%%%%%%%%%%%%%%%%%%%%%%%%%%%%%%%%%%%%%%%%%%
\noindent
{\bf Proposition 2}{
	Considering $m_2<<m_1$, let $\left\{m_1,m_2,\rho(t)\right\} $ and $\left\{{m}_1,{m}'_2,{\rho}'(t)\right\}$ be the surviving and the mortality class size, and the Brier score (BS) before and after balancing respectively. ${\rho}'(t)$ can be approximated as: 
	\begin{eqnarray*}
		{\rho}'(t)= \rho(t)\bigg(\dfrac{m_1+m_2}{{m}_1+{m}'_2}\bigg) \left\{\dfrac{({m}'_2-d_0)e^{-2\hat{H}'_M(t)}+d_0e^{-2{\hat{H}_C(t)}}+{m}_1(1-e^{-{\hat{H}_C(t)}})^2 }{(m_2-d_0)e^{-2\hat{H}_M(t)}+d_0 e^{-2\hat{H}_C(t)}+{m}_1(1-e^{-\hat{H}_C(t)})^2 }\right\}
	\label{IBS_improv}
	\end{eqnarray*}
where $d_0$ is the maximum number of unique death samples present in the leaf nodes.
}\label{proposition}

\noindent
{\bf Proof}. Survival function calculated at a time $t$ for the mortality node $M$, $\hat{S}_{M}(t)$ and the censored node, $\hat{S}_{C}(t)$ are given as follows: 
\begin{eqnarray*}
	\hat{S}_{M}(t)&=& e^{-\hat{H}_M(t)}\\
	\hat{S}_{C}(t)&=& e^{-\hat{H}_C(t)}
\end{eqnarray*}
The prediction error of the nodes are then defined in terms of the expected Brier score (refer to \textbf{subsection 2.2.2}) are given as: 
\begin{eqnarray*}\label{IBS}
\rho(t)= E(\mathcal{Y}_i(t)-\hat{S}_i(t))^2
\end{eqnarray*}
where, $\tilde{\mathcal{Y}}_i= \mathbbm{1}_{{T}_i>t}$ is the actual survival status of individual $i$ at time $t$ and $\hat{S}_i(t)$ is the predicted survival. Further, the predicted survival for individual $i$ is the survival estimator for its leaf node. Given the best survival node split as defined above, BS score calculated for the unbalanced data, $m_1$ and $m_2$ can be represented as: 
\begin{eqnarray*}
	\rho(t)&=&\dfrac{(m_2-d_0)(0-\hat{S}_{M}(t))^2+d_0(0-\hat{S}_{C}(t))^2+ m_1(1-\hat{S}_{C}(t))^2}{m_1+m_2}\\
     &=&\dfrac{(m_2-d_0)e^{-2\hat{H}_M(t)}+d_0e^{-2\hat{H}_C(t)}+m_1(1-e^{-\hat{H}_C(t)})^2 }{m_1+m_2}
\end{eqnarray*}

Now, let us again consider $m_2<<m_1$ and after balancing, let the class proportion be ${m}'_2({m}'_2 \geq m_2)$  (with fixed $m_1$ and $d_0$), the balanced Brier Score can then be given as: 
\begin{eqnarray*}
	{\rho}'(t)=\dfrac{({m}'_2-d_0)e^{-\hat{H}'_M(t)}+d_0e^{-2\hat{H}_C(t)}+{m}_1(1-e^{-\hat{H}_C(t)})^2 }{{m}_1+{m}'_2}
\end{eqnarray*}
Hence the ratio of $\tilde{\rho}(t)$ and $\rho(t)$ can be represented as: 

\begin{eqnarray}
\dfrac{{\rho}'(t)}{\rho(t)}=\bigg(\dfrac{m_1+m_2}{m_1+{m}'_2}\bigg) \left\{\dfrac{({m}'_2-d_0)e^{-2\hat{H}'_M(t)}+d_0e^{-2\hat{H}_C(t)}+m_1(1-e^{-\hat{H}_C(t)})^2 }{(m_2-d_0)e^{-2\hat{H}_M(t)}+d_0 e^{-2\hat{H}_C(t)}+{m}_1(1-e^{-\hat{H}_C(t)})^2 }\right\}
\label{IBS_proposition}
\end{eqnarray}
\hfill\BlackBox\\

%%%%%%%%%%%%%%%%%%%%%%%%%%%%%%%%%%%%%%%%%%%%%%%%%%%%%%%%%%%%%%%%%%%%%%%%%%%%%%%%%%%%%%%%
%% Corollary 1
%%%%%%%%%%%%%%%%%%%%%%%%%%%%%%%%%%%%%%%%%%%%%%%%%%%%%%%%%%%%%%%%%%%%%%%%%%%%%%%%%%%%%%%%%

\noindent
{\bf Corollary 1}{
	Let $\left\lbrace \rho(t), {\rho}'(t)\right\rbrace$  be the Brier scores before and after balancing the class sizes, then ${\rho}'(t)<\rho(t)$.
}\label{cor1}\\

\noindent
{\bf Proof}. Since ${m}'_2>m_2$, we know that $\big(\frac{m_1+m_2}{{m}_1+{m}'_2}\big)<1$. Now, let $f(m_2)=(m_2-d_0)e^{-2H_M(t)}$, showing that $f(m_2)$ is a decreasing function of $m_2$ would suffice to prove \href{cor1}{Corollary 1}. We perform first order differentiation by parts of $f(m_2)$ with respect to $m_2$ which results in: 
\begin{eqnarray}
\frac{df(m_2)}{dm_2}=e^{-2\hat{H}_M(t)}\big(1-2(m_2-d_0)\frac{d\hat{H}_M(t)}{dm_2}\big)
\label{diff}
\end{eqnarray}

We demonstrate Equation~\ref{diff} using $\hat{H}_M(t^{*})$ from Equation~\ref{hazardmortality}, this differentiation is given as follows:

\begin{align*}
\frac{df(m_2)}{dm_2}={}& e^{-{2(y-1)\gamma+\sum_{i=2}^{y}\psi_0(i)}}\big(1-2y\frac{d(2(y-1)\gamma+\psi_0(2)+...\psi_0(m_2+d_0))}{dm_2}\big)\\
={}& e^{-{1.154(y-1)+\sum_{i=2}^{y}\psi_0(i)}}(1-2y(0.577+\psi_1(2)+...+\psi_1(y)))
\end{align*}
Here, $y=(m_2-d_0)$ and $\gamma=0.577$. Further, $\psi_1$ is the Trigamma function~\citep{abramowitz1965handbook} which is positive for non-negative number. Clearly, with exponential and Trigamma function being positive, ${df(m_2)}/{dm_2}<0$. Now that we have established $f(m_2)$ is a decreasing function, for ${m}'_2>m_2$, the right hand side of Equation~\ref{IBS_proposition} becomes less than 1 and hence ${\rho}'(t)<\rho(t)$.
\hfill\BlackBox \\
This implies that the prediction of RSF improves after balancing. For $m_1<<m_2$ similar result holds.\\

\vskip 0.2in
\bibliography{BRSF_arXiv2}

\end{document}